\documentclass[10pt,twocolumn,letterpaper]{article}

\usepackage{iccv}
\usepackage{times}
\usepackage{epsfig}
\usepackage{graphicx}
\usepackage{amsmath}
\usepackage{amssymb}

\usepackage{blindtext}
\usepackage[table]{xcolor}
\usepackage{multirow}
\usepackage{subfig}
\usepackage{makecell}
\usepackage{pifont}

\usepackage{booktabs}
\usepackage{overpic}
\usepackage{authblk}
\usepackage{tikz}

\newcommand\copyrighttextfinal{%
	
	\scriptsize\copyright\ 2023 IEEE. Personal use of this material is permitted. Permission from IEEE must be obtained for all other uses, in any current or future media, including reprinting/republishing this material for advertising or promotional purposes, creating new collective works, for resale or redistribution to servers or lists, or reuse of any copyrighted component of this work in other works.}%
\newcommand\copyrightnotice{%
	
	\begin{tikzpicture}[remember picture,overlay]%
		
		\node[anchor=south,yshift=10pt] at (current page.south) {{\parbox{\dimexpr\textwidth-\fboxsep-\fboxrule\relax}{\copyrighttextfinal}}};%
	\end{tikzpicture}%
	
}

\usepackage[pagebackref=true,breaklinks=true,letterpaper=true,colorlinks,bookmarks=false]{hyperref}

\iccvfinalcopy 

\def\iccvPaperID{3053} 

\ificcvfinal\fi

\begin{document}

\title{Out-of-Distribution Detection for Monocular Depth Estimation}

\author[1]{Julia Hornauer}
\author[1]{Adrian Holzbock}
\author[2]{Vasileios Belagiannis}
\affil[1]{Ulm University, Germany \\ {\tt\small $\{$first.last$\}$@uni-ulm.de}}
\affil[2]{Friedrich-Alexander-Universität Erlangen-Nürnberg, Germany \\ {\tt\small vasileios.belagiannis@fau.de}}

\twocolumn[{%
\renewcommand\twocolumn[1][]{#1}%
\maketitle
\begin{center}
   \centering
   \vspace{-0.2in}
   \begin{overpic}[width=0.9\textwidth]{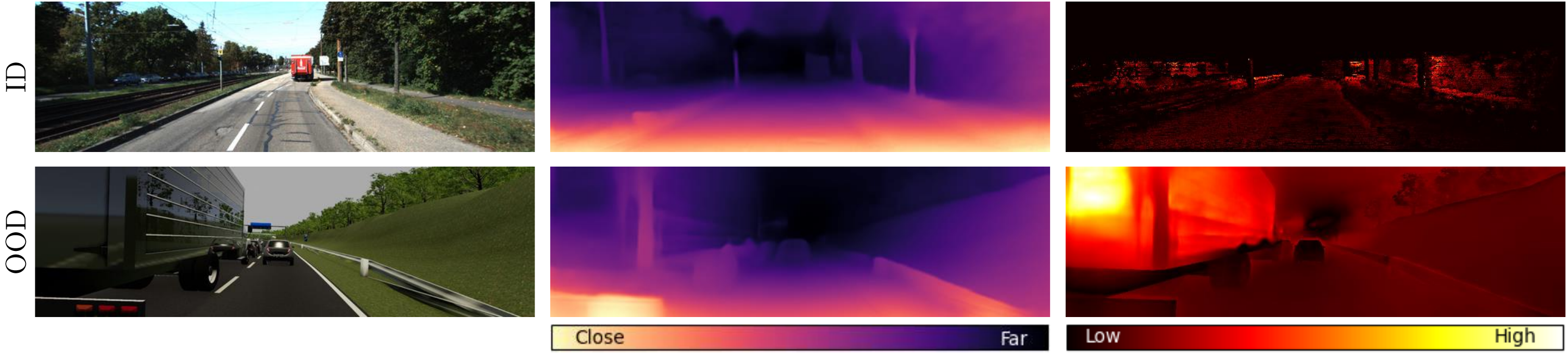}
   \end{overpic}
   \captionof{figure}{Depth prediction (center) and absolute relative error (right) from a model trained on KITTI for images from KITTI (top, left; in-distribution: ID) and virtual KITTI (bottom, left; out-of-distribution: OOD): The depth prediction for virtual KITTI, which is not represented in the training distribution, is incorrect; therefore, the error is too high.}
   \label{fig:teaser}
\end{center}
}]

\maketitle

\ificcvfinal\thispagestyle{empty}\fi

\begin{abstract}
In monocular depth estimation, uncertainty estimation approaches mainly target the data uncertainty introduced by image noise. In contrast to prior work, we address the uncertainty due to lack of knowledge, which is relevant for the detection of data not represented by the training distribution, the so-called out-of-distribution (OOD) data. Motivated by anomaly detection, we propose to detect OOD images from an encoder-decoder depth estimation model based on the reconstruction error. Given the features extracted with the fixed depth encoder, we train an image decoder for image reconstruction using only in-distribution data. Consequently, OOD images result in a high reconstruction error, which we use to distinguish between in- and out-of-distribution samples. 
We built our experiments on the standard NYU Depth V2 and KITTI benchmarks as in-distribution data. Our post hoc method performs astonishingly well on different models and outperforms existing uncertainty estimation approaches without modifying the trained encoder-decoder depth estimation model.
\end{abstract}

\section{Introduction}
\copyrightnotice
Despite the astonishing performance of deep neural networks in monocular depth estimation, image irregularities, e.g., reflections or occlusions, as well as unknown objects or completely unknown environments, lead to incorrect depth predictions. 
Figure~\ref{fig:teaser} shows example predictions (center) from a depth estimation model trained on KITTI~\cite{Geiger2013IJRR} for test images from the same dataset (top, left) as well as virtual KITTI~\cite{cabon2020vkitti2} (bottom, left).
Since virtual KITTI consists of synthetically generated data, the test image is not included in the training data and presents an unknown environment.
The error plot (right) corresponding to the depth prediction (center) shows that the depth estimate for the truck, whose surface has reflections, is incorrectly estimated to be too high. Therefore, the truck's position is incorrect based on the depth estimates, highlighting the importance of identifying inputs for which the depth predictions are unreliable. This is a crucial aspect for the usage in safety-critical applications such as automated driving or robotics.

The erroneous predictions, in general, can be due to different types of uncertainty. The main distinction is between data uncertainty (aleatoric), which results from noise in the data, and model uncertainty (epistemic), which is due to lack of knowledge and therefore addressable with more diverse training data \cite{Kendall2017WhatUD}. Recent works~\cite{Hornauer2022GradientbasedUF,Kendall2017WhatUD,Poggi2020OnTU,Upadhyay2022BayesCapBI} already target uncertainty estimation for depth estimation in the context of deep neural networks. Moreover, they evaluate whether the uncertainty corresponds to the actual error in the test data drawn from the same distribution as the training data. However, the current evaluation protocol only addresses the aleatoric uncertainty but not the epistemic uncertainty. Since epistemic uncertainty results from the lack of knowledge, it is relevant to detecting data not represented by the training distribution, so-called out-of-distribution (OOD) data. 
While OOD detection is an active research area in image classification~\cite{hornauer2023heatmap,Hsu2020GeneralizedOD,liu20_energy,Sun2021ReActOD}, it is rarely explored for complex and high-dimensional tasks like monocular depth estimation. Therefore, we specifically address the detection of OOD samples for depth estimates. 
Motivated by the image anomaly detection approaches that rely on the reconstruction error~\cite{adversarial_ae,ae_survey} of autoencoders to detect anomalous data, we propose to detect OOD inputs from depth estimation models also with the reconstruction error. 

In fact, we train a decoder to reconstruct the input image from the features of an already trained and therefore fixed depth estimation model. Since we train the decoder in a \textit{post hoc} fashion, the depth estimation performance of the underlying model is not affected. More precisely, our image decoder takes the features extracted with the depth encoder of the fixed depth estimation model as input and is trained to predict the original input image. 
To develop our decoder, we rely on a similar model architecture as the depth decoder.
As a result, our method is applicable to various depth estimation models and independent of some specific architecture, i.e., whether the depth estimation model is fully convolutional or transformer-based does not play any role. During inference, the reconstruction error between the predicted and input images serves as the OOD detection score. Our extensive experiments with the OOD detection evaluation protocol show promising results for our method compared to existing uncertainty estimation approaches. 

Our contributions can be summarized as follows: 
    To the best of our knowledge, we are the first to propose an OOD detection method for monocular depth estimation. For this, we do not use OOD examples but rely only on in-distribution training data. 
    In our simple and effective method, we present a second model, represented by an image decoder, to learn image reconstruction post hoc with features extracted by the depth encoder as input. We leverage the reconstruction error as our OOD detection score. 
    We introduce an epistemic uncertainty evaluation protocol that addresses OOD detection with extensive experiments using the standard benchmarks NYU Depth V2~\cite{SilbermanECCV12} and KITTI~\cite{Geiger2013IJRR} as in-distribution data and up to three different OOD datasets. Models and code are publicly available.\footnote{\url{https://github.com/jhornauer/mde_ood}}

\section{Related Work}
\paragraph{Monocular Depth Estimation}
The first neural networks for monocular depth estimation are designed to use information from local and global features~\cite{Eigen2014DepthMP} or as a fully-convolutional architecture~\cite{Laina2016DeeperDP}. To reduce the need for expensive ground truth generation, further works~\cite{Godard2017UnsupervisedMD,Yang2018DeepVS} train their models using stereo pair supervision to leverage the scene geometry. In~\cite{Zhou2017UnsupervisedLO,monodepth2}, the photometric loss is designed to require only monocular sequences, which is more difficult since camera position and scale are unknown and therefore learned together with the depth prediction. 
With the recent success of attention-based transformer architectures in vision tasks, Ranftl et al.~\cite{Ranftl2021VisionTF} are the first to propose using vision transformers~\cite{dosovitskiy2020vit} instead of convolutional backbones to make use of a global receptive field in the encoder.
Bhat et al.~\cite{Bhat2020AdaBinsDE} phrase depth estimation as a classification-regression task where they predict the depth as a linear combination of discretized depth values and probabilities. 
In~\cite{Yuan2022NeuralWF}, the decoder uses energy modeling and multi-head attention to obtain depth predictions. 
Recently, Agarwal et al.~\cite{Agarwal_2023_WACV} propose attention-based skip connections between encoder and decoder. While previous transformer approaches are trained with ground truth supervision, Zhao et al.~\cite{monovit} introduce a transformer architecture for self-supervised monocular depth estimation. In this paper, we perform OOD detection for already trained convolutional and transformer-based depth estimation models. 

\paragraph{Depth Estimation Uncertainty}
In general, a distinction is made between aleatoric and epistemic uncertainty. The first one targets the noise inherent in the data, such as occlusions. In contrast, the latter is the model uncertainty targeting the uncertainty arising from missing knowledge, which is reducible by leveraging more diverse training data and, therefore, relevant for OOD detection~\cite{Kendall2017WhatUD}. More precisely, OOD samples are usually either unknown objects or data represented in a different form, e.g., synthetic images compared to real images \cite{Hsu2020GeneralizedOD}. Epistemic uncertainty estimation approaches place a distribution over the model weights, while aleatoric uncertainty estimation approaches assume a prior distribution over the model output~\cite{Kendall2017WhatUD}.  

A straightforward method to obtain the aleatoric uncertainty with the distribution over two outputs is post-processing as proposed by Godard et al.~\cite{Godard2017UnsupervisedMD}. In~\cite{Hornauer2022GradientbasedUF}, gradients of the model are extracted with this post-processing step to improve uncertainty estimation by using information from output space and gradient space.
Another predictive method infers a prior distribution over the model output by log-likelihood maximization \cite{Nix1994EstimatingTM}. In~\cite{Upadhyay2022BayesCapBI}, the prior output distribution is learned with a second model from a fixed and already trained image-to-image translation model. 
In contrast, we train a second decoder in a post hoc fashion to reconstruct the input image from the features of a depth encoder and thus learn its output distribution. 
Two well-known epistemic uncertainty estimation approaches are Deep Ensembles~\cite{Lakshminarayanan2017SimpleAS} and Monte Carlo Dropout~\cite{mcdropout}. While the first approach generates a distribution over the model weights by training $N$ models with different initialization, the second approach performs $N$ inferences with the same model and dropout~\cite{dropout} applied. To reduce the effort of training $N$ models, Snapshot Ensembles~\cite{HuangSnapshot2017} use cyclic learning rate scheduling. 

Existing uncertainty estimation approaches for depth estimation~\cite{Hornauer2022GradientbasedUF, Poggi2020OnTU} mainly target the aleatoric uncertainty by evaluating whether the predicted uncertainty corresponds to the actual error on the in-distribution test set.
In contrast, we target epistemic uncertainty by proposing an evaluation protocol for detecting OOD inputs with extensive experiments using multiple OOD datasets and different depth estimation models. In this context, we assess the performance of existing uncertainty estimation approaches of both uncertainty types for OOD detection while proposing a new OOD detection method.

\paragraph{Out-of-Distribution Detection}
Many works explore OOD detection for classification~\cite{hendrycks17_baseline,hornauer2023heatmap,liu20_energy}. While the first baseline is the built-in maximum softmax score~\cite{hendrycks17_baseline}, different approaches propose improvements by temperature scaling~\cite{Liang2018EnhancingTR}, the energy function~\cite{liu20_energy} or the group-based softmax~\cite{Huang_2021_CVPR}. 
Another line of research not only relies on in-distribution data but introduces real~\cite{hendrycks2018deep} or synthetic~\cite{lee2018training,Zhang_2023_WACV} outliers during training. In this work, on the other hand, we do not use outliers for OOD detection and only rely on the available in-distribution training data. 
Furthermore, the OOD detection score must not be defined in the output space but can also rely on the weight space~\cite{Sun2021DICELS,Huang2021OnTI}, activation space~\cite{lee18_mahalanobis,Sun2021ReActOD,Dong_2022_CVPR} or both, activation and output space~\cite{Wang_2022_CVPR}.
Cao and Zhang~\cite{Cao_2022_CVPR} directly target the factorization of the different types of uncertainty, namely data, model, and distribution uncertainty, and apply the latter to OOD detection.
Moreover, Wang and Aitchison \cite{wang2022bayesian} use the finding that many uncertainty estimation approaches that predict aleatoric uncertainty~\cite{hendrycks17_baseline,hendrycks2018deep,lee2018training} are effective for OOD detection. Therefore, they integrate aleatoric uncertainty in Bayesian approaches. This work also considers existing aleatoric uncertainty estimation approaches for OOD detection but for monocular depth estimation. 
In contrast, Hendrycks et al.~\cite{NEURIPS2019_a2b15837} leverage representation learning by predicting image rotations, whereas Pei et al.~\cite{Pei2021OutofdistributionDW} detect OOD data with a boundary-aware discriminator. 
Zhou et al.~\cite{Zhou2022RethinkingRA} apply autoencoders for OOD detection. Because it is known that autoencoders also generalize well to unknown inputs, their approach focuses on a maximally compressed latent space while preserving the reconstruction ability. Yang et al.~\cite{Yang2022OutofDistributionDW} synthesize images for OOD detection from the feature space of a classifier with input masking. Similarly, we leverage the reconstruction loss for OOD detection but train a decoder based on the feature space of an already trained depth estimation model for image reconstruction. Recent works expand OOD detection to the more complex tasks object detection~\cite{du2022vos,Du2022UnknownAwareOD,Li2022OutofDistributionIL} and semantic segmentation~\cite{Chan2020EntropyMA,Jung2021StandardizedML,Hendrycks2022ScalingOD}. 
Nevertheless, OOD detection for computationally intensive regression tasks such as monocular depth estimation is still underexplored. 

\section{Out-of-Distribution Detection}
Consider a deep neural network $\mathbf{\hat{d}}=f(\mathbf{x}; \theta)$, paramterized by $\theta$, trained with images $\mathbf{x} \in \mathbb{R}^{w\times h \times 3}$ to predict pixel-wise depth values $\mathbf{\hat{d}} \in \mathbb{R}^{w\times h \times 1}$, where $w$ and $h$ are width and height. The model is trained with the dataset $\mathcal{D}=\{\mathbf{x}_{i},\mathbf{d}_{i}\}_{i=1}^{|\mathcal{D}|}$ consisting of images $\mathbf{x}$ and corresponding depth $\mathbf{d} \in \mathbb{R}^{w \times h \times 1}$. The dataset is drawn from the training distribution $\mathcal{P}_{in}(\mathbf{x},\mathbf{d})$, referred to as \textit{in-distribution}. 
During inference, the model cannot just be exposed to images $\textbf{x}_{in}$ originating from $\mathcal{P}_{in}$ but also to images $\mathbf{x}_{out}$. 
The so-called \textit{out-of-distribution} inputs $\mathbf{x}_{out}$ are drawn from a different distribution $\mathcal{P}_{out}(\mathbf{x}, \mathbf{d})$, which is different from the training distribution $\mathcal{P}_{in}(\mathbf{x}, \mathbf{d})$. Such OOD inputs can either contain unknown objects or have a non-semantic shift~\cite{Hsu2020GeneralizedOD}, e.g., synthetic compared to real data. This work aims to detect such OOD samples $\mathbf{x}_{out}$. 

Given an already trained depth estimation model, we propose to detect OOD inputs without modifying the model parameters. This is referred to as post hoc OOD detection.  
Most depth estimation models are built with a feature encoder $\mathbf{z}=\psi(\mathbf{x}, \theta_{\psi})$ and a depth decoder $\mathbf{\hat{d}}=\phi(\mathbf{z};\theta_{\phi})$, where the features $\mathbf{z}=\{\textbf{z}_j\}_{j=1}^{M}$ are extracted from $M$ different encoder layers. 
We assume to have access to the features $\mathbf{z}$ of the encoder-decoder depth estimation model, but we keep its model weights fixed. Therefore, our method is independent of the underlying training strategy of the depth estimation model. Moreover, the depth estimation model can be trained either in a supervised manner with the depth ground truth $\mathbf{d}$ or in a self-supervised manner without $\mathbf{d}$. 

\subsection{Training Approach}
\begin{figure}
    \centering
    \includegraphics[width=0.9\linewidth]{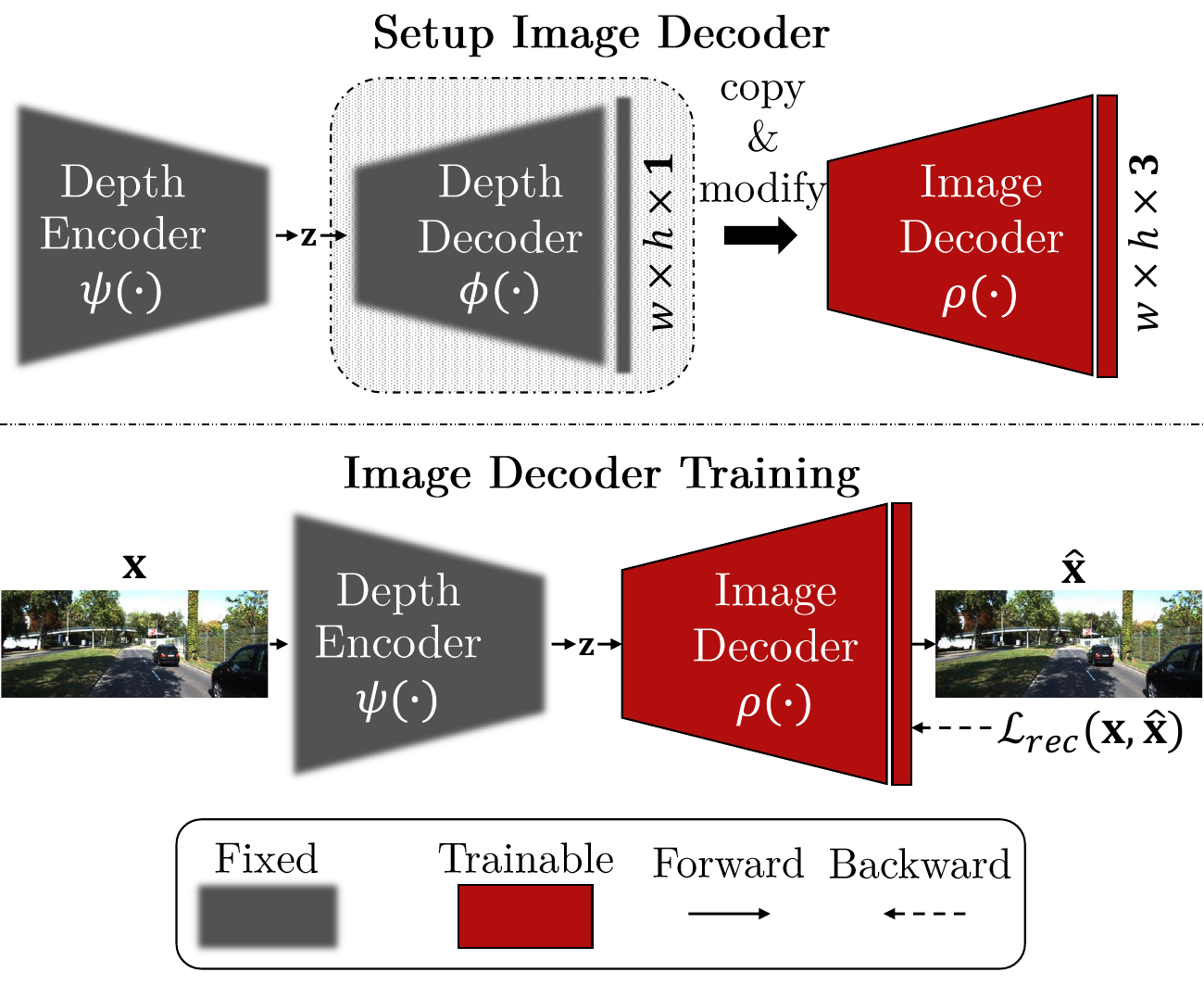}
    \caption{Given a trained and fixed depth estimation model, we use a similar model architecture as the depth decoder by copying and modifying it to use as an image decoder.
    During training, the image decoder receives the encoder features as input and is optimized to reconstruct the input images only using in-distribution data. Note that we only back-propagate through the image decoder.} 
    \label{fig:train}
\end{figure}

Inspired by anomaly detection~\cite{ae_survey}, where an autoencoder is optimized to learn the representation of data defined as \textit{normal}, we propose to learn a representation of the training distribution with a second decoder, namely an image decoder. The image decoder $\mathbf{\hat{x}}=\rho(\mathbf{z};\theta_{\rho})$, parameterized by $\theta_{\rho}$, is trained to reconstruct the original input images $\textbf{x}$ from the features $\mathbf{z}$ extracted with the already trained and fixed depth encoder. The reconstructed images $\mathbf{\hat{x}} \in \mathbb{R}^{w \times h \times 3}$ have the same dimension as the original images $\mathbf{x}$.  
Since we train the image decoder only on in-distribution data, we assume access to the images $\mathbf{x}_{in}$ drawn from the in-distribution training dataset $\mathcal{D}$.
Similar to anomaly detection~\cite{ae_survey}, we define the in-distribution images $\mathbf{x}_{in}$ as \textit{normal} and assume that the decoder reconstructs them well. In contrast, the decoder should not be able to reconstruct OOD images since the image decoder did not observe them during optimization.  
The poor reconstruction of OOD images is also demonstrated in Figure~\ref{fig:inference}. In the case of the OOD input, the reconstructed image $\mathbf{\hat{x}}$ shows a color shift in comparison to the original input $\mathbf{x}$. The difference between the original image $\mathbf{x}$ and the reconstructed image $\mathbf{\hat{x}}$ can then be used to define an OOD detection score.  

\paragraph{Decoder Training}
Figure~\ref{fig:train} illustrates an overview of our approach. First, we make a copy of the architecture of the depth decoder $\phi(\cdot)$ and modify it to use it as our image decoder $\rho(\cdot)$. 
By choosing the image decoder to have a similar architecture as the depth decoder, we make our method applicable to various depth estimation models independent of the architecture, i.e., whether the model is fully convolutional or transformer-based.
Note that the image decoder is trained from scratch and does not reuse the weights of the depth decoder.
During the image decoder training, we replace the depth estimation loss function with the reconstruction loss: 
\begin{equation}
    \mathcal{L}_{rec}(\mathbf{\hat{x}}, \mathbf{x})= \mathbb{E}_{\mathbf{x} \sim \mathcal{D}}| \mathbf{\hat{x}} - \mathbf{x} |.
\end{equation}
Thus, the image decoder learns to reconstruct the original input image $\mathbf{x}$ from the features $\mathbf{z}$ instead of predicting the pixel-wise depth. Note that we only back-propagate through the image decoder for the gradient update. 
This makes our approach simple and feasible for different kinds of depth estimation models. In our experiments, we demonstrate the effectiveness of our approach for convolutional and transformer-based depth estimation models (see Sec.~\ref{sec:results}). 

\subsection{Out-of-Distribution Detection Score}
\begin{figure*}
\centering
    \includegraphics[width=0.9\linewidth]{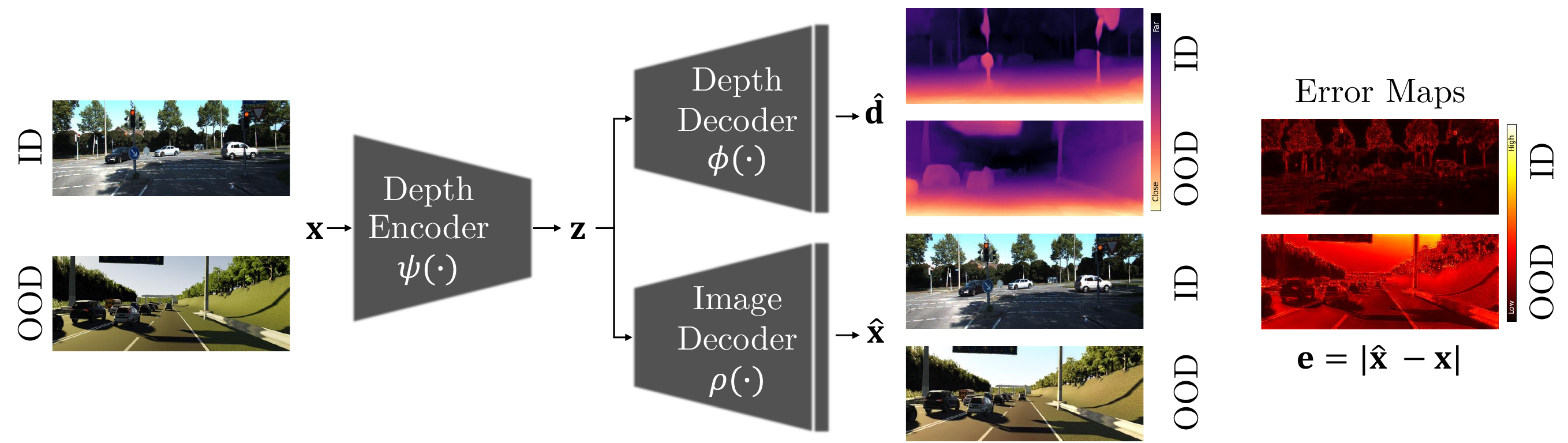}
    \caption{During inference, the image decoder $\rho(\cdot)$ is used next to the depth decoder $\phi(\cdot)$. Since the image decoder $\rho(\cdot)$ is only trained on in-distribution images, it cannot correctly reconstruct the input images in the case of OOD data. Because of this reason, the reconstruction error between the input image $\mathbf{x}$ and the reconstructed image $\mathbf{\hat{x}}$ is used to calculate error maps $\mathbf{e}$, which in turn are used to detect OOD inputs.}
    \label{fig:inference}
\end{figure*}
In Figure~\ref{fig:inference}, the inference process is illustrated. The input image $\mathbf{x}$ is passed through the depth estimation encoder and decoder as well as our image decoder to obtain the depth prediction $\mathbf{\hat{d}}$ and the input reconstruction $\mathbf{\hat{x}}$. Afterward, we leverage the predicted images $\mathbf{\hat{x}}$ to detect OOD inputs. First, we use the reconstruction error $\mathbf{e}_{abs} = |\mathbf{\hat{x}} - \mathbf{x}|$ to to obtain the pixel-wise absolute error $\mathbf{e}_{abs} \in \mathbb{R}^{w \times h \times 3}$ between the input image $\mathbf{x}$ and the predicted image $\mathbf{\hat{x}}$.
Then, we apply max-pooling in the channel dimensions to obtain the error map $\mathbf{e} \in \mathbb{R}^{w \times h \times 1}$. Finally, we use the error map to calculate the OOD detection score:
\begin{equation} \label{ood_score}
    \text{score}(\mathbf{e}) = \begin{cases} 
          1 & \frac{1}{w \cdot h} \sum_{k=1}^{w \cdot h} \mathbf{e}_{k} \leq \tau \\
          0 & \text{otherwise}.
       \end{cases}
\end{equation}
 In-distribution samples are set to $1$, while OOD samples are set to $0$. The threshold $\tau$ is used to separate in- from out-of-distribution data. It can be chosen manually, but all metrics are threshold independent in our evaluation. Whereas in-distribution inputs should result in a low reconstruction error, OOD inputs should lead to high reconstruction errors. This, in turn, makes OOD inputs distinguishable from in-distribution data (see Figure~\ref{fig:inference}).

\section{Experiments}
We evaluate the OOD detection performance with our epistemic uncertainty evaluation protocol based on two standard depth estimation benchmarks as in-distribution data and up to three OOD datasets and different model architectures. First, we explain the evaluation protocol, including in- and out-of-distribution datasets and metrics. In addition, we describe the used models and implementation details. Then, we show the OOD detection results compared to related work and conduct an ablation study on the choice to train the image decoder in a post hoc manner. 

\subsection{Experimental Setup}
\paragraph{Datasets}
We propose to evaluate the OOD detection on two different setups where the depth estimation model is trained on NYU Depth V2~\cite{SilbermanECCV12} or KITTI~\cite{Geiger2013IJRR} as an in-distribution dataset. 
NYU is an indoor depth estimation dataset recorded with a resolution of $480 \times 640$ at 494 scenes. We use the official train/test split with a maximum depth of 10 meters. 
For NYU as in-distribution data, we propose to use Places365~\cite{Zhou2018PlacesA1} (Places) as OOD data, which consists of scenes captured in different indoor and outdoor environments. We choose images from the outdoor scenes \textit{butte}, \textit{cliff}, \textit{corn field}, \textit{desert road}, \textit{harbor}, and \textit{highway}. 
Since OOD inputs are rare compared to in-distribution data in real-world scenarios, we choose a smaller number of OOD images than the number of in-distribution images. We select 50 images per category for Places to obtain 300 OOD images, while the NYU test set contains 654 samples. 

KITTI, by contrast, is an autonomous driving dataset with an average resolution of $375 \times 1242$ captured at 61 scenes in Germany. We use the Eigen split~\cite{Eigen2014DepthMP} with the maximum depth set to 80 meters for evaluation. In \cite{Upadhyay2022BayesCapBI}, a small OOD evaluation is conducted with KITTI as in-distribution and Places365~\cite{Zhou2018PlacesA1} and India Driving~\cite{Varma2018IDDAD} as OOD. In contrast to this evaluation, we suggest using two different in-distribution datasets and treating the OOD datasets as separate settings instead of combining them. Also, as previously mentioned, instead of using all OOD test samples, we limit the number of OOD images to 300 to accommodate real-world scenarios. The KITTI test set, on the other hand, has twice as many images, with 652. 
For KITTI as in-distribution, we choose three different OOD datasets instead of only two, namely Places~\cite{Zhou2018PlacesA1}, India Driving~\cite{Varma2018IDDAD} (India) and virtual KITTI~\cite{cabon2020vkitti2} (vKITTI). The three databases cover different types of OOD data. India encompasses driving scenes on Indian roads, whereas vKITTI consists of synthetic driving scenarios. 
For Places, we select 50 samples from the indoor scenes \textit{art gallery}, \textit{bathroom}, \textit{dining room}, \textit{home office}, \textit{hospital room} and \textit{kitchen}. For India and vKITTI, we randomly choose the 300 test samples. In all constellations, we use the entire in-distribution test set and the selected OOD samples of the respective OOD test set. Finally, all images are resized to the training image resolution for the OOD evaluation.
\paragraph{Evaluation Metrics}
Consistent with the literature on OOD detection for classification~\cite{liu20_energy, lee18_mahalanobis, Huang2021OnTI}, we adopt the standard OOD detection metrics.
We use the area under the receiver operating curve (AUROC), the area under the precision-recall curve with in-distribution as positives (AUPRS), as well as with OOD as positives (AUPRE) and the false positive rate at 95\% true positive rate (FPR95). Note that all metrics are threshold-independent. 
\paragraph{Models}
Compared to recent work~\cite{Upadhyay2022BayesCapBI,Poggi2020OnTU,Hornauer2022GradientbasedUF}, we do not only evaluate on Monodepth2~\cite{monodepth2} but also on two recently published transformer-based models Pixelformer~\cite{Agarwal_2023_WACV} and MonoViT~\cite{monovit}. 
In the case of NYU, Monodepth2 and Pixelformer are trained in a supervised manner.
While Monodepth2 is trained with an image resolution of $224 \times 288$, Pixelformer operates on the original image resolution of $480 \times 640$. In the case of KITTI, we also use Monodepth2 and the recent transformer architecture MonoViT. Both models are trained self-supervised with monocular supervision and an image resolution of $192 \times 640$.
\paragraph{Comparison to Related Work}
We compare our approach to the aleatoric uncertainty estimation approaches post-processing (Post~\cite{Godard2017UnsupervisedMD}), log-likelihood maximization (Log~\cite{Klodt2018SupervisingTN,Nix1994EstimatingTM}) and BayesCap (BCap~\cite{Upadhyay2022BayesCapBI}). Moreover, we also evaluate against the epistemic uncertainty estimation approach Monte Carlo Dropout (Drop~\cite{mcdropout}). Since they all predict pixel-wise uncertainty, we use the mean uncertainty over the pixels as the OOD detection score, similar to~\cite{Upadhyay2022BayesCapBI}. Note that Post can be applied to all models without training, while BCap is also trained post hoc. Log and Drop, on the other hand, must be integrated into the training strategy of the depth estimation model by modifying the training loss function or inserting dropout layers, respectively.
\paragraph{Implementation Details} 
For Monodepth2, Pixelformer, and MonoViT, we use the code provided by \cite{monodepth2}, \cite{Agarwal_2023_WACV}, and \cite{monovit}, respectively.
For the setup and training of our image reconstruction decoders, we modify the architecture of the respective depth decoder and apply the same optimizer and hyperparameters. In the case of Monodepth2 and MonoViT, we remove the Sigmoid activation function after the last convolution and change the output channels of the last convolution from $1$ to $3$. Since Pixelformer phrases depth estimation as a classification-regression task, we remove the Bin Center Predictor and only rely on the upsampling path with the Skip Attention Modules. In addition, we remove the Softmax activation function in the disparity head and add a convolution that reduces the output channels to $3$.
For Log, the depth estimation models are trained with the loss functions as purposed in~\cite{Klodt2018SupervisingTN,Nix1994EstimatingTM}. Moreover, for Drop, dropout is applied at different decoder stages with the dropout probability $0.2$. Furthermore, the BCap models are trained with the implementations provided by~\cite{Upadhyay2022BayesCapBI}. We provide further details in the supplementary material.  

\subsection{Out-of-Distribution Detection Results}\label{sec:results}
\paragraph{Results on NYU}
\begin{table}[]
    \centering
    \begin{tabular}{@{\hskip1pt}lc@{\hskip3pt}@{\hskip3pt}c@{\hskip3pt}@{\hskip3pt}c@{\hskip3pt}@{\hskip3pt}c@{\hskip1pt}}
    \toprule
    Method & AUROC$\uparrow$ & AUPRS$\uparrow$ & AUPRE$\uparrow$ & FPR95$\downarrow$ \\
    \midrule
    Post~\cite{Godard2017UnsupervisedMD} & 56.42 & 70.22 & 41.22 & 88.33 \\
    Log~\cite{Nix1994EstimatingTM} & \textbf{77.32} & \textbf{87.51} & \underline{64.73} & \underline{68.00} \\
    BCap~\cite{Upadhyay2022BayesCapBI} & 34.95 & 57.42 & 32.88 & 89.67 \\ 
    Drop~\cite{mcdropout} & 49.50 & 67.15 & 34.77 & 90.67 \\
    Ours & \underline{74.84} & \underline{83.50} & \textbf{65.14} & \textbf{65.33} \\
    \bottomrule
    \end{tabular}
    \caption{Evaluation of Monodepth2 trained on NYU with Places as OOD.} 
    \label{tab:nyu-monodepth}
\end{table}
Table~\ref{tab:nyu-monodepth} lists the OOD detection results for Monodepth2 trained on NYU as in-distribution data and Places as OOD dataset. 
The best AUROC and AUPRS results are obtained with Log, while our method obtains the highest AUPR and FPR95. Since the depth estimation model is trained with depth ground truth, Log has the advantage that it learns to estimate the actual error. Therefore, it is a good representation of the training distribution.
However, the loss function for model training must be changed to the log-likelihood maximization loss, and thus the depth estimation performance is influenced. 
Although Drop targets epistemic uncertainty, it has the worst OOD detection performance in this experiment.
\begin{table}[]
    \centering
    \begin{tabular}{@{\hskip1pt}lc@{\hskip3pt}@{\hskip3pt}c@{\hskip3pt}@{\hskip3pt}c@{\hskip3pt}@{\hskip3pt}c@{\hskip1pt}}
    \toprule
    Method & AUROC$\uparrow$ & AUPRS$\uparrow$ & AUPRE$\uparrow$ & FPR95$\downarrow$ \\
    \midrule
    Post~\cite{Godard2017UnsupervisedMD} & 79.12 & 88.46 & 63.38 & 67.67 \\
    Log~\cite{Nix1994EstimatingTM} & 40.68 & 61.73 & 26.38 & 97.33 \\
    BCap~\cite{Upadhyay2022BayesCapBI} & 39.68 & 63.35 & 26.35 & 97.33 \\ 
    Drop~\cite{mcdropout} & \underline{88.13} & \underline{93.10} & \underline{81.99} & \underline{46.00} \\
    Ours & \textbf{93.18} & \textbf{95.62} & \textbf{90.88} & \textbf{21.67} \\
    \bottomrule
    \end{tabular}
    \caption{Evaluation of Pixelformer trained on NYU with Places as OOD.}
    \label{tab:nyu-pixelformer}
\end{table}

Table~\ref{tab:nyu-pixelformer} states the results on the same dataset constellation but for the recent transformer-based architecture Pixelformer. Here, our method obtains the best OOD detection outcome, with FPR95, in particular, being comparatively low for our approach, followed by Drop and Post. Compared to the previous experiment with Monodepth2, Log performs worse, while the OOD detection results of Post, Drop, and our method are improved overall.   
NYU is a large dataset with 50K images comprising different indoor environments, which makes OOD detection challenging. Pixelformer has the advantage of using global attention and significantly more model parameters, which already results in a better depth estimation performance compared to Monodepth2. The results indicate that the global attention and larger model capacity enable Pixelformer to learn a richer feature representation. This helps to obtain a good representation of the in-distribution data.

\paragraph{Results on KITTI}
For KITTI as in-distribution data, the OOD detection performance is stated separately for the three OOD datasets Places, India, and vKITTI. The results for Monodepth2 trained on KITTI are provided in Table~\ref{tab:monodepth-kittim}. Overall, the OOD detection performance of the uncertainty estimation approaches varies for the different OOD datasets. It underscores the need to evaluate the different OOD datasets as separate domains rather than as a mixture of OOD data. 
Our approach consistently outperforms all uncertainty estimation approaches in all three constellations. Nevertheless, Post also achieves good results for Places, which is the simplest setup since it does not include traffic scenarios but indoor environments. Although Post likewise obtains the second-best results for the other two setups, the OOD detection performance is still significantly worse compared to our approach. Similar to the findings in~\cite{Gustafsson2019EvaluatingSB}, our experiments show that the epistemic uncertainty detection approach Drop is unsuitable for OOD detection in high-dimensional problems such as monocular depth estimation. The results show that the existing uncertainty approaches are insufficient to detect OOD inputs, and a new method, explicitly targeting OOD detection, is required.  
\begin{table}[]
    \centering
    \begin{tabular}{@{\hskip1pt}l@{\hskip3pt}@{\hskip3pt}lc@{\hskip3pt}@{\hskip3pt}c@{\hskip3pt}@{\hskip3pt}c@{\hskip3pt}@{\hskip3pt}c@{\hskip1pt}}
    \toprule
     & Method & AUROC$\uparrow$ & AUPRS$\uparrow$ & AUPRE$\uparrow$ & FPR95$\downarrow$ \\
     \midrule
     \multirow{5}{*}{\rotatebox{90}{Places}} & Post~\cite{Godard2017UnsupervisedMD} & \underline{97.43} & \underline{98.68} & \underline{95.38} & \underline{11.67} \\
     & Log~\cite{Klodt2018SupervisingTN} & 33.95 & 57.97 & 25.13 & 96.67 \\
     & BCap~\cite{Upadhyay2022BayesCapBI} & 71.68 & 79.43 & 59.86 & 69.67 \\
     & Drop~\cite{mcdropout} & 25.47 & 56.52 & 21.07 & 99.67\\
     & Ours  & \textbf{99.38} & \textbf{99.67} & \textbf{98.95} & \textbf{3.33}
     \\
     \midrule
     \multirow{5}{*}{\rotatebox{90}{India}} & Post~\cite{Godard2017UnsupervisedMD} & \underline{59.61} & \underline{77.31} & \underline{36.96} & 94.33 \\
     & Log~\cite{Klodt2018SupervisingTN} & 41.08 & 62.03 & 30.96 & \underline{91.00} \\
     & BCap~\cite{Upadhyay2022BayesCapBI} & 40.85 & 64.42 & 25.61 & 99.00 \\
     & Drop~\cite{mcdropout} & 53.36 & 70.91 & 34.74 & 92.67 \\
     & Ours & \textbf{89.29} & \textbf{92.30} & \textbf{87.89} & \textbf{25.00} \\
    \midrule
    \multirow{5}{*}{\rotatebox{90}{vKITTI}} & Post~\cite{Godard2017UnsupervisedMD} & \underline{75.98} & \underline{87.17} & \underline{55.47} & \underline{81.67} \\
     & Log~\cite{Klodt2018SupervisingTN} & 25.97 & 55.15 & 21.38 & 98.67 \\
     & BCap~\cite{Upadhyay2022BayesCapBI} & 47.93 & 67.64 & 30.40 & 96.00 \\
     & Drop~\cite{mcdropout} & 36.54 & 60.16 & 24.71 & 97.67 \\
     & Ours & \textbf{99.40} & \textbf{99.68} & \textbf{99.04} & \textbf{2.00} \\
    \bottomrule
    \end{tabular}
    \caption{Evaluation of Monodepth2 trained on KITTI with Places, India and vKITTI as OOD.}
    \label{tab:monodepth-kittim}
\end{table}

Table~\ref{tab:monovit-kittim} states the results for MonoViT trained on KITTI. The OOD detection metrics show a similar tendency as for Monodepth2. Again, our approach outperforms all uncertainty estimation approaches. Post obtains the second-best outcomes for Places and vKITTI, whereas Drop is the second-best method in the case of India. Especially for the FPR95 metric, a critical indication to determine a threshold $\tau$ in practice, there is a large gap between the uncertainty estimation approaches and our reconstruction-based OOD detection method. 
\begin{table}[]
    \centering
    \begin{tabular}{@{\hskip1pt}l@{\hskip3pt}@{\hskip3pt}lc@{\hskip3pt}@{\hskip3pt}c@{\hskip3pt}@{\hskip3pt}c@{\hskip3pt}@{\hskip3pt}c@{\hskip1pt}}
    \toprule
     & Method & AUROC$\uparrow$ & AUPRS$\uparrow$ & AUPRE$\uparrow$ & FPR95$\downarrow$ \\
     \midrule
     \multirow{5}{*}{\rotatebox{90}{Places}}& Post~\cite{Godard2017UnsupervisedMD} & \underline{94.16} & \underline{97.24} & \underline{83.39} & \underline{35.00} \\
     & Log~\cite{Klodt2018SupervisingTN} & 22.54 & 52.44 & 21.82 &  96.67 \\
     & BCap~\cite{Upadhyay2022BayesCapBI} & 52.25 & 69.76 & 31.12 & 99.67 \\
     & Drop~\cite{mcdropout} & 68.67 & 83.38 & 48.98 & 85.67 \\
     & Ours & \textbf{99.70} & \textbf{99.84} & \textbf{99.52} & \textbf{1.33} \\
     \midrule
     \multirow{5}{*}{\rotatebox{90}{India}}& Post~\cite{Godard2017UnsupervisedMD} & 46.53 & 67.65 & 28.19 & 99.67  \\
     & Log~\cite{Klodt2018SupervisingTN} & 35.25 & 57.84 & 25.15 & 97.67 \\
     & BCap~\cite{Upadhyay2022BayesCapBI} & 51.12 & 69.47 & 30.33 & 100.00 \\
     & Drop~\cite{mcdropout} & \underline{64.20} & \underline{77.34} & \underline{46.30} & \underline{82.67} \\
     & Ours & \textbf{95.95} & \textbf{97.40} & \textbf{94.84} & \textbf{14.67} \\
    \midrule
    \multirow{5}{*}{\rotatebox{90}{vKITTI}} & Post~\cite{Godard2017UnsupervisedMD} & \underline{68.03} & \underline{81.65} & \underline{43.34} & 96.00 \\
     & Log~\cite{Klodt2018SupervisingTN} & 25.44 & 54.02 & 21.21 & 99.33 \\
     & BCap~\cite{Upadhyay2022BayesCapBI} & 56.07 & 72.88 & 33.88 & 100.00 \\
     & Drop~\cite{mcdropout} & 51.40 & 69.49 & 34.51 & \underline{92.00} \\
     & Ours & \textbf{97.48} & \textbf{98.64} & \textbf{96.19} & \textbf{9.00} \\
    \bottomrule
    \end{tabular}
    \caption{Evaluation of MonoViT trained on KITTI with Places, India and vKITTI as OOD.} 
    \label{tab:monovit-kittim}
\end{table}
\paragraph{Visual Results}
\begin{figure}
    \centering
    \begin{tabular}{@{\hskip0pt}c@{\hskip2pt}@{\hskip2pt}c@{\hskip2pt}@{\hskip2pt}c@{\hskip0pt}}
        \rotatebox[origin=lt]{90}{RGB} & \includegraphics[width=0.45\linewidth]{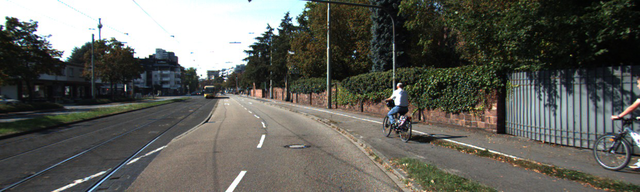} &  \includegraphics[width=0.45\linewidth]{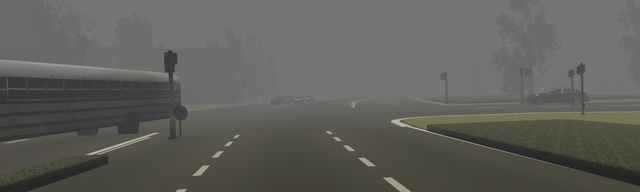} \\
        \rotatebox[origin=lt]{90}{Error} & \includegraphics[width=0.45\linewidth]{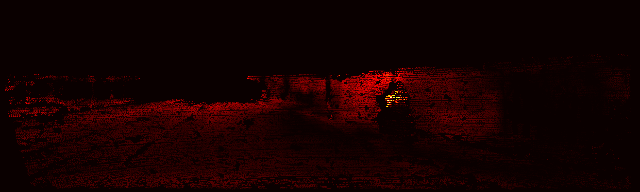} & \includegraphics[width=0.45\linewidth]{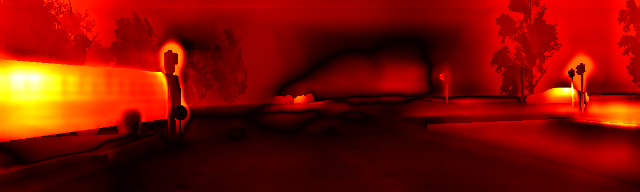} \\
        \rotatebox[origin=lt]{90}{Post} & \includegraphics[width=0.45\linewidth]{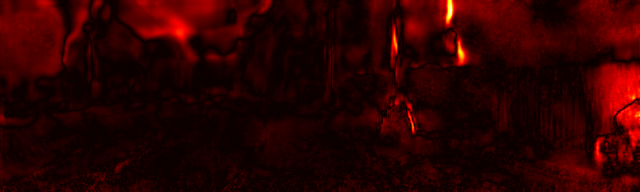} & \includegraphics[width=0.45\linewidth]{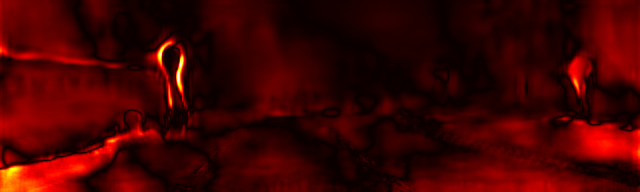} \\ 
        \rotatebox[origin=lt]{90}{Log} & \includegraphics[width=0.45\linewidth]{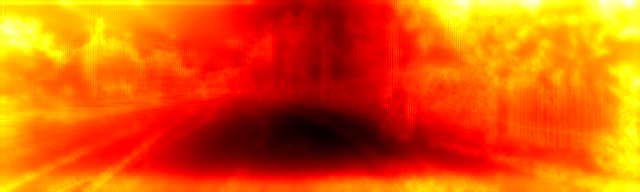} & \includegraphics[width=0.45\linewidth]{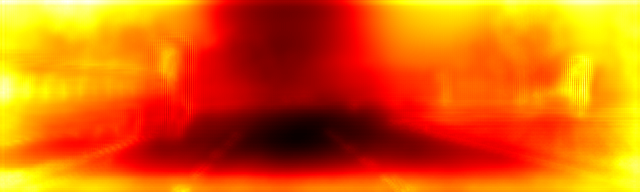} \\
        \rotatebox[origin=lt]{90}{BCap} & \includegraphics[width=0.45\linewidth]{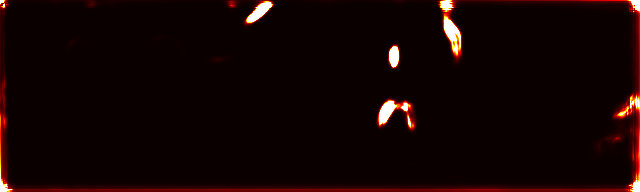} & \includegraphics[width=0.45\linewidth]{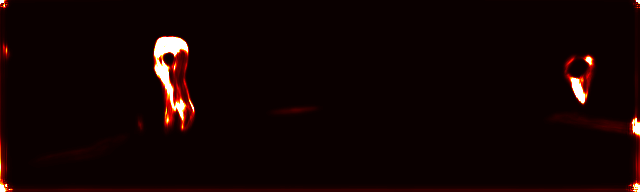} \\
        \rotatebox[origin=lt]{90}{Drop} & \includegraphics[width=0.45\linewidth]{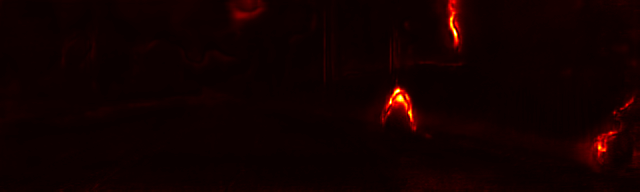} &
        \includegraphics[width=0.45\linewidth]{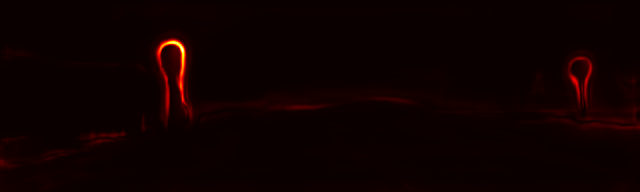} \\
        \vspace{-5pt}
        \rotatebox[origin=lt]{90}{Ours} & \includegraphics[width=0.45\linewidth]{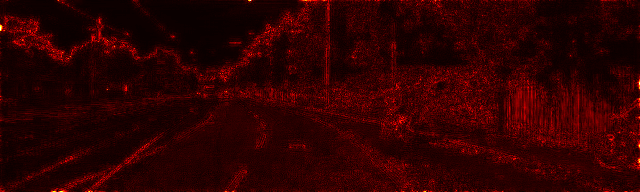} & \includegraphics[width=0.45\linewidth]{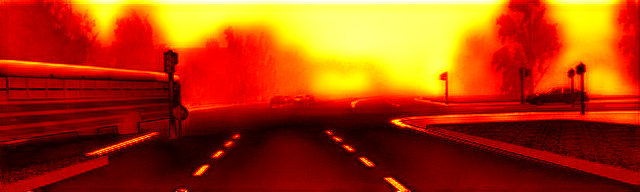} \\
        \vspace{-3pt}
         & \includegraphics[width=0.46\linewidth]{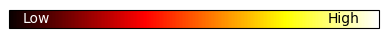} & \includegraphics[width=0.46\linewidth]{images/colorbars/hot_colorbar.png} \\
        & KITTI & vKITTI \\
    \end{tabular}
    \caption{The first row shows test samples (RGB) from the in-distribution dataset KITTI (left) and the OOD dataset vKITTI (right) with corresponding absolute relative error in the second row (Error) for MonoViT. The pixel-wise uncertainties of the approaches Post, Log, Drop, and BCap are shown in the following rows. The last row displays the error map of our reconstruction-based OOD detection method.}
    \label{fig:visu_examples}
\end{figure}
Figure~\ref{fig:visu_examples} illustrates test samples from KITTI and vKITTI (first row) with corresponding absolute relative error (second row), followed by the uncertainty maps of the different approaches in the next rows and the error map of our approach (last row). It is visible that the test sample from vKITTI is synthetic and, therefore, from a different domain. In particular, the bus's depth on the image's left side is misjudged. The bus would, therefore, not have been detected according to the depth map. The yellow and light red colors in the error map (Error) also emphasize this.
In contrast, the prediction error for the in-distribution test sample is much smaller. The uncertainty maps from Post, Log, BCap, and Drop show similar uncertainty estimates for the in- and out-of-distribution test samples. Notably, the OOD input's uncertainty is similar to the estimated uncertainty for the in-distribution image. For example, the uncertainty of Drop is low for both cases, while the uncertainty of Log is high for both inputs. Therefore, the OOD input cannot be differentiated from in-distribution data based on uncertainty estimates. It demonstrates that more than the uncertainty estimation methods, regardless of the aleatoric or epistemic approach, are needed to estimate the uncertainty of OOD inputs. Furthermore, the uncertainty for OOD inputs should be high, especially for high prediction errors. It underlines that an OOD detection method is required to filter such inputs. 
The error maps of our method show a low reconstruction error for the in-distribution image, whereas the error for the OOD image is high. It shows that in high-dimensional problems, such as depth estimation, image reconstruction shows astonishing performance in OOD detection. The improved performance is explainable by the decoder, which is trained with the depth features and captures the different color schemes of the in-distribution data. More precisely, it can reconstruct the contours of the OOD inputs but not the correct color schemes. This color shift results in a high reconstruction error, making our method a suitable OOD detection approach. 
\subsection{Depth Estimation Results on vKITTI}\label{sec:results_vkitti}
Table~\ref{tab:depth-vkitti} reports the depth estimation results for Monodepth2 and MonoViT trained on KITTI and evaluated on KITTI and the OOD test set vKITTI. The results are provided using the standard metrics~\cite{monovit,monodepth2,Agarwal_2023_WACV} absolute relative error (Abs Rel), root mean squared error (RMSE) and accuracy $\delta_{1}$ ($\delta < 1.25$).
When comparing the results, it is evident that the depth estimates in all three metrics are significantly worse. It highlights the need for detecting OOD inputs for robust depth estimation in real-world applications. 
\begin{table}[]
    \centering
    \begin{tabular}{crccc}
    \toprule
       Model & Dataset & Abs Rel$\downarrow$ & RMSE$\downarrow$ & $\delta_{1}$$\uparrow$ \\
    \midrule
     \multirow{2}{*}{MD} & KITTI & 0.090 & 3.942 & 0.914 \\
     & vKITTI & 0.305 & 15.787 & 0.511 \\
     \midrule
     \multirow{2}{*}{MV} & KITTI & 0.079 & 3.466 & 0.934 \\
     & vKITTI & 0.257 & 14.683 & 0.563  \\
    \bottomrule 
    \end{tabular}
    \caption{Depth estimation results for Monodepth2 (MD) and MonViT (MV) trained on KITTI and evaluated on KITTI and the OOD dataset vKITTI.} 
    \label{tab:depth-vkitti}
\end{table}

\subsection{Ablation Studies}
In this section, we evaluate the image decoder training strategy. Instead of training the image decoder post hoc, we ablate the OOD detection performance if the image decoder is optimized simultaneously with the depth estimation model (Sim) or if the encoder and the image decoder are trained from scratch as autoencoder (AE). For Monodepth2 and MonoViT, we provide the depth estimation performance of the original models (Depth) compared to training the image decoder simultaneously with the depth estimation model (Sim) in Table~\ref{tab:depth-kitti}. The results demonstrate that the depth estimation performance decreases if we optimize the image decoder with the depth estimation model. 
In Table~\ref{tab:sim-train}, the OOD detection performance averaged over the three datasets Places, India, and vKITTI is given for both models. The results show that the autoencoder leads to the worst OOD detection performance.
Analyzing the reconstructed images, the autoencoder reconstructs all images well regardless of in-distribution or OOD. 
The proposed image decoder, by contrast, only reconstructs the in-distribution images well. 
This implies that the task-specific features provided by the trained depth estimation model help the image decoder to focus on the in-distribution data due to the fixed and therefore constrained feature representation. The autoencoder, on the other hand, extracts general features with the aim of reconstruction, leading to generalization beyond in-distribution data.
Furthermore, the OOD detection results are similar when the image decoder is trained post hoc or with the depth estimation model. While concurrent training slightly improves the performance of Monodepth2, it is the opposite for MonoViT. However, post hoc training of image reconstruction has the advantage that the performance of depth estimation is not affected by the training of the image decoder and is thus independent of OOD detection. 
\begin{table}[]
    \centering
    \begin{tabular}{ccccc}
    \toprule
       Model & Method & Abs Rel$\downarrow$ & RMSE$\downarrow$ & $\delta_{1}$$\uparrow$ \\
    \midrule
    \multirow{2}{*}{MD} & Depth & 0.090 & 3.942 & 0.914 \\
        & Sim & 0.101 & 4.260 & 0.892 \\
    \midrule
    \multirow{2}{*}{MV} & Depth & 0.079 & 3.466 & 0.934 \\
     & Sim & 0.085 & 3.696 &  0.922 \\
    \bottomrule
    \end{tabular}
    \caption{Depth estimation results of Monodepth2 (MD) and MonoViT (MV) trained on KITTI without the image decoder (Depth) or together with the image decoder (Sim).}
    \label{tab:depth-kitti}
\end{table}
\begin{table}[]
    \centering
    \begin{tabular}{@{\hskip1pt}c@{\hskip3pt}@{\hskip3pt}c@{\hskip3pt}@{\hskip3pt}c@{\hskip3pt}@{\hskip3pt}c@{\hskip3pt}@{\hskip3pt}c@{\hskip3pt}@{\hskip3pt}c@{\hskip1pt}}
    \toprule
      Model & Method & AUROC$\uparrow$ & AUPRS$\uparrow$ & AUPRE$\uparrow$ & FPR95$\downarrow$ \\
      \midrule
      \multirow{3}{*}{MD} & AE & 52.17 & 69.13 & 38.62 & 87.33 \\
      & Sim & \textbf{98.57} & \textbf{99.32} & \textbf{97.07} & \textbf{6.78} \\
      & Ours & 96.02 & 97.22 & 95.29 & 10.11 \\
      \midrule
      \multirow{3}{*}{MV} & AE & 35.04 & 57.22 & 28.35 & 94.33 \\ 
      & Sim & 94.20 & 95.15 & 93.21 & 14.89 \\
      & Ours & \textbf{97.71} & \textbf{98.63} & \textbf{96.85} & \textbf{8.44} \\
      \bottomrule
    \end{tabular}
    \caption{OOD evaluation of our method compared to training the image decoder together with the depth estimation model (Sim) or using an autoencoder (AE) for Monodepth2 (MD) and MonoViT (MV). We train the models on KITTI as in-distribution data and average the OOD detection results over the three datasets Places, India, and vKITTI.}
    \label{tab:sim-train}
\end{table}
\section{Conclusion}
Motivated by anomaly detection, we present a simple and effective OOD detection approach for monocular depth estimation. Starting from an already trained encoder-decoder depth estimation model, we train an image decoder post hoc for input reconstruction using features extracted by the fixed depth encoder. Designing our image decoder architecture similar to the depth decoder makes our approach applicable to different models regardless of their model architecture, i.e., whether they are fully convolutional or transformer-based. Compared to existing uncertainty estimation approaches, our method, which relies on the reconstruction error as the OOD detection score, shows impressive performance on the epistemic uncertainty evaluation protocol with the benchmarks NYU Depth V2 and KITTI as the in-distribution data and different OOD datasets.

{\small
\bibliographystyle{ieee_fullname}
\bibliography{iccv_2023}

\begin{thebibliography}{10}\itemsep=-1pt

\bibitem{Agarwal_2023_WACV}
Ashutosh Agarwal and Chetan Arora.
\newblock Attention attention everywhere: Monocular depth prediction with skip
  attention.
\newblock In {\em Proceedings of the IEEE/CVF Winter Conference on Applications
  of Computer Vision (WACV)}, pages 5861--5870, January 2023.

\bibitem{adversarial_ae}
Laura Beggel, Michael Pfeiffer, and Bernd Bischl.
\newblock Robust anomaly detection in images using adversarial autoencoders.
\newblock In Ulf Brefeld, Elisa Fromont, Andreas Hotho, Arno Knobbe, Marloes
  Maathuis, and C{\'e}line Robardet, editors, {\em Machine Learning and
  Knowledge Discovery in Databases}, pages 206--222, Cham, 2020. Springer
  International Publishing.

\bibitem{Bhat2020AdaBinsDE}
S. Bhat, Ibraheem Alhashim, and Peter Wonka.
\newblock Adabins: Depth estimation using adaptive bins.
\newblock {\em 2021 IEEE/CVF Conference on Computer Vision and Pattern
  Recognition (CVPR)}, pages 4008--4017, 2020.

\bibitem{cabon2020vkitti2}
Yohann Cabon, Naila Murray, and Martin Humenberger.
\newblock Virtual kitti 2.
\newblock 2020.

\bibitem{Cao_2022_CVPR}
Senqi Cao and Zhongfei Zhang.
\newblock Deep hybrid models for out-of-distribution detection.
\newblock In {\em Proceedings of the IEEE/CVF Conference on Computer Vision and
  Pattern Recognition (CVPR)}, pages 4733--4743, June 2022.

\bibitem{Chan2020EntropyMA}
Robin Chan, Matthias Rottmann, and Hanno Gottschalk.
\newblock Entropy maximization and meta classification for out-of-distribution
  detection in semantic segmentation.
\newblock {\em 2021 IEEE/CVF International Conference on Computer Vision
  (ICCV)}, pages 5108--5117, 2020.

\bibitem{ae_survey}
Varun Chandola, Arindam Banerjee, and Vipin Kumar.
\newblock Anomaly detection: A survey.
\newblock {\em ACM computing surveys (CSUR)}, 41(3):1--58, 2009.

\bibitem{Dong_2022_CVPR}
Xin Dong, Junfeng Guo, Ang Li, Wei-Te Ting, Cong Liu, and H.T. Kung.
\newblock Neural mean discrepancy for efficient out-of-distribution detection.
\newblock In {\em Proceedings of the IEEE/CVF Conference on Computer Vision and
  Pattern Recognition (CVPR)}, pages 19217--19227, June 2022.

\bibitem{dosovitskiy2020vit}
Alexey Dosovitskiy, Lucas Beyer, Alexander Kolesnikov, Dirk Weissenborn,
  Xiaohua Zhai, Thomas Unterthiner, Mostafa Dehghani, Matthias Minderer, Georg
  Heigold, Sylvain Gelly, Jakob Uszkoreit, and Neil Houlsby.
\newblock An image is worth 16x16 words: Transformers for image recognition at
  scale.
\newblock In {\em International Conference on Learning Representations}, 2021.

\bibitem{Du2022UnknownAwareOD}
Xuefeng Du, Xin Wang, Gabriel Gozum, and Yixuan Li.
\newblock Unknown-aware object detection: Learning what you don't know from
  videos in the wild.
\newblock {\em 2022 IEEE/CVF Conference on Computer Vision and Pattern
  Recognition (CVPR)}, pages 13668--13678, 2022.

\bibitem{du2022vos}
Xuefeng Du, Zhaoning Wang, Mu Cai, and Yixuan Li.
\newblock Vos: Learning what you don’t know by virtual outlier synthesis.
\newblock {\em Proceedings of the International Conference on Learning
  Representations}, 2022.

\bibitem{Eigen2014DepthMP}
David Eigen, Christian Puhrsch, and Rob Fergus.
\newblock Depth map prediction from a single image using a multi-scale deep
  network.
\newblock In {\em NIPS}, 2014.

\bibitem{mcdropout}
Yarin Gal and Zoubin Ghahramani.
\newblock Dropout as a bayesian approximation: Representing model uncertainty
  in deep learning.
\newblock {\em Proceedings of The 33rd International Conference on Machine
  Learning}, 06 2015.

\bibitem{Geiger2013IJRR}
Andreas Geiger, Philip Lenz, Christoph Stiller, and Raquel Urtasun.
\newblock Vision meets robotics: The kitti dataset.
\newblock {\em International Journal of Robotics Research (IJRR)}, 2013.

\bibitem{Godard2017UnsupervisedMD}
Cl{\'e}ment Godard, Oisin~Mac Aodha, and Gabriel~J. Brostow.
\newblock Unsupervised monocular depth estimation with left-right consistency.
\newblock {\em 2017 IEEE Conference on Computer Vision and Pattern Recognition
  (CVPR)}, pages 6602--6611, 2017.

\bibitem{monodepth2}
Cl{\'e}ment Godard, Oisin~Mac Aodha, and Gabriel~J. Brostow.
\newblock Digging into self-supervised monocular depth estimation.
\newblock {\em 2019 IEEE/CVF International Conference on Computer Vision
  (ICCV)}, pages 3827--3837, 2019.

\bibitem{Gustafsson2019EvaluatingSB}
Fredrik~K. Gustafsson, Martin Danelljan, and Thomas~Bo Sch{\"o}n.
\newblock Evaluating scalable bayesian deep learning methods for robust
  computer vision.
\newblock {\em 2020 IEEE/CVF Conference on Computer Vision and Pattern
  Recognition Workshops (CVPRW)}, pages 1289--1298, 2019.

\bibitem{Hendrycks2022ScalingOD}
Dan Hendrycks, Steven Basart, Mantas Mazeika, Mohammadreza Mostajabi, Jacob
  Steinhardt, and Dawn~Xiaodong Song.
\newblock Scaling out-of-distribution detection for real-world settings.
\newblock In {\em International Conference on Machine Learning}, 2022.

\bibitem{hendrycks17_baseline}
Dan Hendrycks and Kevin Gimpel.
\newblock A baseline for detecting misclassified and out-of-distribution
  examples in neural network.
\newblock In {\em International Conference on Learning Representations}, 2017.

\bibitem{hendrycks2018deep}
Dan Hendrycks, Mantas Mazeika, and Thomas Dietterich.
\newblock Deep anomaly detection with outlier exposure.
\newblock In {\em International Conference on Learning Representations}, 2019.

\bibitem{NEURIPS2019_a2b15837}
Dan Hendrycks, Mantas Mazeika, Saurav Kadavath, and Dawn Song.
\newblock Using self-supervised learning can improve model robustness and
  uncertainty.
\newblock In H. Wallach, H. Larochelle, A. Beygelzimer, F. d\textquotesingle
  Alch\'{e}-Buc, E. Fox, and R. Garnett, editors, {\em Advances in Neural
  Information Processing Systems}, volume~32. Curran Associates, Inc., 2019.

\bibitem{Hornauer2022GradientbasedUF}
Julia Hornauer and Vasileios Belagiannis.
\newblock Gradient-based uncertainty for monocular depth estimation.
\newblock In {\em European Conference on Computer Vision}, 2022.

\bibitem{hornauer2023heatmap}
Julia Hornauer and Vasileios Belagiannis.
\newblock Heatmap-based out-of-distribution detection.
\newblock In {\em Proceedings of the IEEE/CVF Winter Conference on Applications
  of Computer Vision}, pages 2603--2612, 2023.

\bibitem{Hsu2020GeneralizedOD}
Yen-Chang Hsu, Yilin Shen, Hongxia Jin, and Zsolt Kira.
\newblock Generalized odin: Detecting out-of-distribution image without
  learning from out-of-distribution data.
\newblock {\em 2020 IEEE/CVF Conference on Computer Vision and Pattern
  Recognition (CVPR)}, pages 10948--10957, 2020.

\bibitem{HuangSnapshot2017}
Gao Huang, Yixuan Li, Geoff Pleiss, Zhuang Liu, John~E. Hopcroft, and Kilian~Q.
  Weinberger.
\newblock Snapshot ensembles: Train 1, get {M} for free.
\newblock In {\em 5th International Conference on Learning Representations,
  {ICLR} 2017, Toulon, France, April 24-26, 2017, Conference Track
  Proceedings}. OpenReview.net, 2017.

\bibitem{Huang2021OnTI}
Rui Huang, Andrew Geng, and Yixuan Li.
\newblock On the importance of gradients for detecting distributional shifts in
  the wild.
\newblock In {\em Neural Information Processing Systems}, 2021.

\bibitem{Huang_2021_CVPR}
Rui Huang and Yixuan Li.
\newblock Mos: Towards scaling out-of-distribution detection for large semantic
  space.
\newblock In {\em Proceedings of the IEEE/CVF Conference on Computer Vision and
  Pattern Recognition (CVPR)}, pages 8710--8719, June 2021.

\bibitem{Jung2021StandardizedML}
Sanghun Jung, Jungsoo Lee, Daehoon Gwak, Sungha Choi, and Jaegul Choo.
\newblock Standardized max logits: A simple yet effective approach for
  identifying unexpected road obstacles in urban-scene segmentation.
\newblock {\em 2021 IEEE/CVF International Conference on Computer Vision
  (ICCV)}, pages 15405--15414, 2021.

\bibitem{Kendall2017WhatUD}
Alex Kendall and Yarin Gal.
\newblock What uncertainties do we need in bayesian deep learning for computer
  vision?
\newblock In {\em NIPS}, 2017.

\bibitem{Klodt2018SupervisingTN}
Maria Klodt and Andrea Vedaldi.
\newblock Supervising the new with the old: Learning sfm from sfm.
\newblock In {\em European Conference on Computer Vision}, 2018.

\bibitem{Laina2016DeeperDP}
Iro Laina, C. Rupprecht, Vasileios Belagiannis, Federico Tombari, and Nassir
  Navab.
\newblock Deeper depth prediction with fully convolutional residual networks.
\newblock {\em 2016 Fourth International Conference on 3D Vision (3DV)}, pages
  239--248, 2016.

\bibitem{Lakshminarayanan2017SimpleAS}
Balaji Lakshminarayanan, Alexander Pritzel, and Charles Blundell.
\newblock Simple and scalable predictive uncertainty estimation using deep
  ensembles.
\newblock In {\em NIPS}, 2017.

\bibitem{lee2018training}
Kimin Lee, Honglak Lee, Kibok Lee, and Jinwoo Shin.
\newblock Training confidence-calibrated classifiers for detecting
  out-of-distribution samples.
\newblock In {\em International Conference on Learning Representations}, 2018.

\bibitem{lee18_mahalanobis}
Kimin Lee, Kibok Lee, Honglak Lee, and Jinwoo Shin.
\newblock A simple unified framework for detecting out-of-distribution samples
  and adversarial attacks.
\newblock In S. Bengio, H. Wallach, H. Larochelle, K. Grauman, N. Cesa-Bianchi,
  and R. Garnett, editors, {\em NIPS}, volume~31. Curran Associates, Inc.,
  2018.

\bibitem{Li2022OutofDistributionIL}
Ruoqi Li, Chongyang Zhang, Hao Zhou, Chao Shi, and Yan Luo.
\newblock Out-of-distribution identification: Let detector tell which i am not
  sure.
\newblock In {\em European Conference on Computer Vision}, 2022.

\bibitem{Liang2018EnhancingTR}
Shiyu Liang, Yixuan Li, and Rayadurgam Srikant.
\newblock Enhancing the reliability of out-of-distribution image detection in
  neural networks.
\newblock {\em International Conference on Learning Representations}, 2018.

\bibitem{liu20_energy}
Weitang Liu, Xiaoyun Wang, John Owens, and Yixuan Li.
\newblock Energy-based out-of-distribution detection.
\newblock In H. Larochelle, M. Ranzato, R. Hadsell, M.~F. Balcan, and H. Lin,
  editors, {\em NIPS}, volume~33, pages 21464--21475. Curran Associates, Inc.,
  2020.

\bibitem{SilbermanECCV12}
Pushmeet~Kohli Nathan~Silberman, Derek~Hoiem and Rob Fergus.
\newblock Indoor segmentation and support inference from rgbd images.
\newblock In {\em European Conference on Computer Vision}, 2012.

\bibitem{Nix1994EstimatingTM}
David~A. Nix and Andreas~S. Weigend.
\newblock Estimating the mean and variance of the target probability
  distribution.
\newblock {\em Proceedings of 1994 IEEE International Conference on Neural
  Networks (ICNN'94)}, 1:55--60 vol.1, 1994.

\bibitem{Pei2021OutofdistributionDW}
Sen Pei, Xin Zhang, Bin Fan, and Gaofeng Meng.
\newblock Out-of-distribution detection with boundary aware learning.
\newblock In {\em European Conference on Computer Vision}, 2021.

\bibitem{Poggi2020OnTU}
Matteo Poggi, Filippo Aleotti, Fabio Tosi, and S. Mattoccia.
\newblock On the uncertainty of self-supervised monocular depth estimation.
\newblock {\em 2020 IEEE/CVF Conference on Computer Vision and Pattern
  Recognition (CVPR)}, pages 3224--3234, 2020.

\bibitem{Ranftl2021VisionTF}
Ren{\'e} Ranftl, Alexey Bochkovskiy, and Vladlen Koltun.
\newblock Vision transformers for dense prediction.
\newblock {\em 2021 IEEE/CVF International Conference on Computer Vision
  (ICCV)}, pages 12159--12168, 2021.

\bibitem{dropout}
Nitish Srivastava, Geoffrey Hinton, Alex Krizhevsky, Ilya Sutskever, and Ruslan
  Salakhutdinov.
\newblock Dropout: A simple way to prevent neural networks from overfitting.
\newblock {\em Journal of Machine Learning Research}, 15(56):1929--1958, 2014.

\bibitem{Sun2021ReActOD}
Yiyou Sun, Chuan Guo, and Yixuan Li.
\newblock React: Out-of-distribution detection with rectified activations.
\newblock In {\em NeurIPS}, 2021.

\bibitem{Sun2021DICELS}
Yiyou Sun and Yixuan Li.
\newblock Dice: Leveraging sparsification for out-of-distribution detection.
\newblock In {\em European Conference on Computer Vision}, 2021.

\bibitem{Upadhyay2022BayesCapBI}
Uddeshya Upadhyay, Shyamgopal Karthik, Yanbei Chen, Massimiliano Mancini, and
  Zeynep Akata.
\newblock Bayescap: Bayesian identity cap for calibrated uncertainty in frozen
  neural networks.
\newblock In {\em European Conference on Computer Vision}, 2022.

\bibitem{Varma2018IDDAD}
G. Varma, A. Subramanian, Anoop~M. Namboodiri, Manmohan Chandraker, and C.~V.
  Jawahar.
\newblock Idd: A dataset for exploring problems of autonomous navigation in
  unconstrained environments.
\newblock {\em 2019 IEEE Winter Conference on Applications of Computer Vision
  (WACV)}, pages 1743--1751, 2018.

\bibitem{Wang_2022_CVPR}
Haoqi Wang, Zhizhong Li, Litong Feng, and Wayne Zhang.
\newblock Vim: Out-of-distribution with virtual-logit matching.
\newblock In {\em Proceedings of the IEEE/CVF Conference on Computer Vision and
  Pattern Recognition (CVPR)}, pages 4921--4930, June 2022.

\bibitem{wang2022bayesian}
Xi Wang and Laurence Aitchison.
\newblock Bayesian {OOD} detection with aleatoric uncertainty and outlier
  exposure.
\newblock In {\em Fourth Symposium on Advances in Approximate Bayesian
  Inference}, 2022.

\bibitem{Yang2018DeepVS}
Nan Yang, Rui Wang, J. St{\"u}ckler, and Daniel Cremers.
\newblock Deep virtual stereo odometry: Leveraging deep depth prediction for
  monocular direct sparse odometry.
\newblock In {\em European Conference on Computer Vision}, 2018.

\bibitem{Yang2022OutofDistributionDW}
Yijun Yang, Ruiyuan Gao, and Qiang Xu.
\newblock Out-of-distribution detection with semantic mismatch under masking.
\newblock In {\em European Conference on Computer Vision}, 2022.

\bibitem{Yuan2022NeuralWF}
Weihao Yuan, Xiaodong Gu, Zuozhuo Dai, Siyu Zhu, and Ping Tan.
\newblock Neural window fully-connected crfs for monocular depth estimation.
\newblock {\em 2022 IEEE/CVF Conference on Computer Vision and Pattern
  Recognition (CVPR)}, pages 3906--3915, 2022.

\bibitem{Zhang_2023_WACV}
Jingyang Zhang, Nathan Inkawhich, Randolph Linderman, Yiran Chen, and Hai Li.
\newblock Mixture outlier exposure: Towards out-of-distribution detection in
  fine-grained environments.
\newblock In {\em Proceedings of the IEEE/CVF Winter Conference on Applications
  of Computer Vision (WACV)}, pages 5531--5540, January 2023.

\bibitem{monovit}
Chaoqiang Zhao, Youmin Zhang, Matteo Poggi, Fabio Tosi, Xianda Guo, Zheng Zhu,
  Guan Huang, Yang Tang, and Stefano Mattoccia.
\newblock Monovit: Self-supervised monocular depth estimation with a vision
  transformer.
\newblock In {\em International Conference on 3D Vision}, 2022.

\bibitem{Zhou2018PlacesA1}
Bolei Zhou, {\`A}gata Lapedriza, Aditya Khosla, Aude Oliva, and Antonio
  Torralba.
\newblock Places: A 10 million image database for scene recognition.
\newblock {\em IEEE Transactions on Pattern Analysis and Machine Intelligence},
  40:1452--1464, 2018.

\bibitem{Zhou2017UnsupervisedLO}
Tinghui Zhou, Matthew~A. Brown, Noah Snavely, and David~G. Lowe.
\newblock Unsupervised learning of depth and ego-motion from video.
\newblock {\em 2017 IEEE Conference on Computer Vision and Pattern Recognition
  (CVPR)}, pages 6612--6619, 2017.

\bibitem{Zhou2022RethinkingRA}
Yibo Zhou.
\newblock Rethinking reconstruction autoencoder-based out-of-distribution
  detection.
\newblock {\em 2022 IEEE/CVF Conference on Computer Vision and Pattern
  Recognition (CVPR)}, pages 7369--7377, 2022.

\end{thebibliography}
}

\end{document}